\documentclass[runningheads]{llncs}

\usepackage{eccv}

\usepackage{eccvabbrv}

\usepackage{graphicx}
\usepackage{booktabs}

\usepackage[accsupp]{axessibility}  %

\usepackage{bm}
\usepackage{amsmath}
\usepackage{amssymb}
\usepackage{nicefrac}
\usepackage{dsfont}
\usepackage{colortbl}
\usepackage{comment}
\usepackage{caption}
\usepackage{wrapfig}

\usepackage[pagebackref,breaklinks,colorlinks,citecolor=eccvblue]{hyperref}

\usepackage{orcidlink}

\renewcommand{\paragraph}[1]{\vspace{1.25mm}\noindent\textbf{#1}}

\definecolor{baselinecolor}{gray}{.9}
\definecolor{darkgreen}{rgb}{0.13, 0.55, 0.13}

\definecolor{mygray}{gray}{0.91}
\definecolor{fontgray}{gray}{0.7}

\definecolor{mygreen}{HTML}{67C055}
\definecolor{orangearrow}{HTML}{ee8641}

\definecolor{myorange}{HTML}{F2A93B}
\definecolor{mypink}{HTML}{D84BEF}

\newcommand{\perfdown}[1]{$\,_{\text{\bf\textcolor{BrickRed}{(#1)}}}$}
\newcommand{\perfup}[1]{$\,_{\text{\bf\textcolor{darkgreen}{(#1)}}}$}

\begin{document}

\title{Labeled Data Selection for Category Discovery}

\author{
Bingchen Zhao\inst{1} \quad
Nico Lang\inst{2} \quad
Serge Belongie\inst{2} \quad
Oisin Mac Aodha\inst{1}
}

\authorrunning{B.~Zhao et al.}

\institute{
$^1$University of Edinburgh \quad $^2$University of Copenhagen
}

\maketitle

\begin{abstract}
Visual category discovery methods aim to find novel categories in unlabeled visual data. 
At training time, a set of labeled and unlabeled images are provided, where the labels correspond to the categories present in the images. 
The labeled data provides guidance during training by indicating what types of visual properties and features are relevant for performing discovery in the unlabeled data. 
As a result, changing the categories present in the labeled set can have a large impact on what is ultimately discovered in the unlabeled set. 
Despite its importance, the impact of labeled data selection has not been explored in the category discovery literature to date. 
We show that changing the labeled data does indeed significantly impact discovery performance. 
Motivated by this, we propose two new approaches for automatically selecting the most suitable labeled data based on the similarity between the labeled and unlabeled data. 
Our observation is that, unlike in conventional supervised transfer learning, the most informative labeled data is neither too similar nor too dissimilar, to the unlabeled categories. 
Our resulting approaches obtain state-of-the-art discovery performance across a range of methods and challenging fine-grained benchmark datasets. 
\end{abstract}

\section{Introduction}
\label{sec:intro}

Given a set of labeled and unlabeled images, the goal of category discovery methods is to automatically group images in the unlabeled data into categories. 
The problem has been explored under several different guises, including settings where the unlabeled data only contains images from novel categories~\cite{han2019learning}, where the unlabeled data can contain both previously seen \emph{and} novel categories~\cite{vaze2022generalized}, and also in the online learning setting~\cite{zhao2023incremental,zhang2022grow}. 
What differentiates discovery from conventional unsupervised clustering~\cite{ji2019invariant,caron2018deep} is the addition of the labeled data.

\begin{figure}[ht]
\centering
\includegraphics[width=\linewidth]{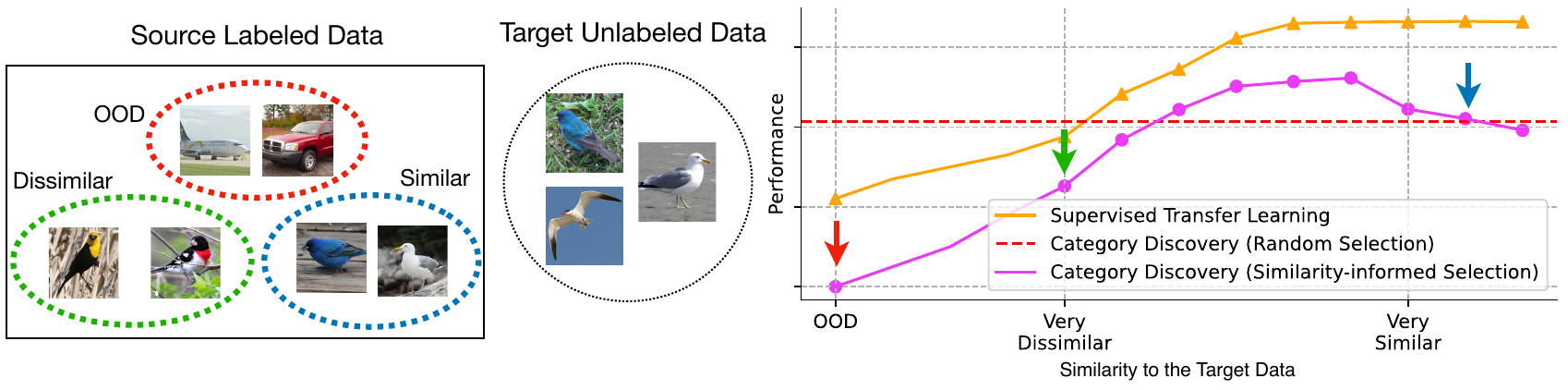}
\vspace{-18pt}
\caption{\textbf{Discovery benefits from selecting a subset of the labeled data.} Category discovery methods rely on labeled source data to provide context for discovering visual concepts in unlabeled target data. 
In the related task of supervised categorization via transfer learning~\cite{Cui_2018_CVPR}, performance on the target data is maximized by selecting the most related concepts from the source data to pre-train the model (\textcolor{myorange}{\bf orange} line). 
We show that discovery methods behave very differently (\textcolor{mypink}{\bf pink} line). Learning representations from source data that is too similar to the target actually hinders discovery. 
Instead, it is preferable to select data that is neither too similar nor too dissimilar. 
The arrows \textcolor{red}{$\downarrow$} \textcolor{green}{$\downarrow$} \textcolor{blue}{$\downarrow$} (right) indicate the locations of the subsets (left) with the same color.
}
\label{fig:intro_fig}
\vspace{-18pt}
\end{figure}

On first appearance, the availability of labeled data may appear to only be a minor difference from traditional unsupervised clustering. 
However, the labeled data plays an important role for discovery algorithms. 
It provides guidance during training regarding the types of visual features that are deemed important and relevant for defining categories in the unlabeled data.  
For example, suppose one was provided with a dataset that contains images captured in nature. 
If the labeled data only contained images of different species of birds, this strongly limits the discovery method to only be able to find `bird-like' concepts in the unlabeled data.   
Similarly, if the labeled data instead only contained images of plant species, it would restrict the method to discover `plant-like' concepts. 

As a result, the role and selection of labeled data is an important question in category discovery. 
However, until now, this problem has been overlooked in the discovery literature, which has instead focused on developing new methods that can be applied to off-the-shelf datasets with already provided labeled and unlabeled splits. 
In the context of supervised classification, \cite{Cui_2018_CVPR} showed that the choice of labeled pre-training source data can have a large impact on downstream fine-tuned target performance. 
Motivated by this, they proposed an approach for selecting a source training set that is most similar to a target domain of interest by measuring the similarity between the source and target domains. 
Their observation is that the ideal source dataset for supervised classification is the one that is most similar to the target domain. 
In this work, we show that this is \emph{not} the case for category discovery. 

In discovery, we demonstrate that the similarity between the labeled categories and the unlabeled data exposes a surprising, and at first perhaps counter-intuitive, relationship.  
Intuitively, if the categories in the labeled data are too \emph{dissimilar} to the ones in the unlabeled data, discovery fails, \ie labeled images of insects do not provide useful cues for discovering species of birds. 
However, if the labeled categories are too \emph{similar} to the unlabeled ones this also dramatically hurts discovery performance, \ie if the labeled and unlabeled datasets both contain very visually similar, but distinct, fine-grained bird categories, discovery methods are unable to separate the two sets of similar categories.  
As a result, there appears to be a `sweet spot' in the middle of both extremes, whereby the labeled images should be related enough to provide useful context, but not too similar as to be overly difficult to separate. 
We show that exploiting this observation to select a subset of labeled categories for training category discovery methods results in significantly enhanced discovery performance (see~\cref{fig:intro_fig}). 

We introduce a new labeled dataset selection framing of the generalized category discovery problem. 
At training time we are presented with an unlabeled target dataset in which we want to discover novel categories. 
In addition, we are given a labeled source dataset where the goal is to automatically select a subset from it to train the discovery algorithm. 
Importantly, this labeled dataset can potentially contain images from diverse categories, many of which may not be relevant to the unlabeled target. 
We propose two new unsupervised selection methods that assign weight to the labeled data during the training process so that labeled data that is too similar or too dissimilar to the target data is down-weighted and thus has less influence on the learned representation. 

We make the following contributions: 
(i) We show that category discovery performance highly depends on the specific labeled data used during training. 
(ii) We propose two new methods for automatically selecting a more effective labeled training dataset to improve discovery performance. 
(iii) We show that our approaches generalize across multiple discovery methods and result in state-of-the-art performance on several challenging fine-grained discovery benchmarks.

\vspace{-10pt}
\section{Related work}
\label{sec:rel_work}

\subsection{Category discovery} 
The task of category discovery centers around the development of algorithms that aim to find novel categories in unlabeled data.  
In addition to unlabeled data, labeled data is also provided at training time which gives information indicating the types of visual concepts that are to be discovered. 
Several different variants of the core category discovery problem have been proposed in the literature. 

In novel category discovery (NCD), the unlabeled data is assumed to not contain any images from the categories in the labeled set~\cite{han2019learning,Han2020automatically,han2021autonovel,zhao2021novel,Fini_2021_ICCV,Zhong_2021_CVPR}. 
This is a overly simplifying assumption, and in the more realistic generalized category discovery (GCD) setting the unlabeled data can contain images from both previously seen and novel categories~\cite{vaze2022generalized,cao22,wen2023parametric,pu2023dynamic,fei2022xcon,zhang2022promptcal,vaze2023no}. 
Both of these discovery paradigms operate under a closed set assumption, \ie there are no novel categories in the test set that are not in the labeled or unlabeled sets.  
On-the-fly category discovery (OCD) attempts to handle this specific challenge~\cite{du2023fly,rastegar2023learn}.
Finally, there is a recent set of methods that explore discovery in the online learning setting~\cite{zhang2022grow,kim2023proxy,zhao2023incremental}. 
Here, the goal is to not only discover new concepts but also not to forget previously seen ones. 

The above methods have primarily focused on methodological improvements and have neglected to investigate the role of training data. 
Theoretical works on category discovery have also emerged of late.  
\cite{chi2022meta} examined the bound for category discovery when the number of examples is limited and~\cite{sun2023a} and~\cite{sun2023and} investigated how labeled data helps the discovery of novel categories from a spectral analysis perspective. 
In this work, we explore the impact of the labeled data in the generalized discovery setting. 
We show that the choice of which labeled data to use can significantly impact downstream discovery performance. 
Based on this, we outline a simple and practical solution for selecting labeled data for category discovery. 
Our results, across a range of recent methods, illustrate that state-of-the-art discovery performance can be obtained by simply training on selected labeled data. 

\vspace{-10pt}
\subsection{Labeled data selection}  
\vspace{-5pt}
The conventional wisdom in supervised categorization is that more data, coupled with larger models, results in improved performance~\cite{sun2017revisiting,dehghani2023scaling}.  
However, not all data is equally useful and commonly used supervised training datasets in vision have required large amounts of manual effort to collect and curate~\cite{russakovsky2015imagenet,lin2014microsoft,van2015building,van2018inaturalist}. 
Existing work has shown that both supervised~\cite{xu2023demystifying,fang2023data} and self-supervised~\cite{oquab2023dinov2} approaches can benefit from semi-automated methods for data filtering when constructing training datasets. 
There is also a set of works that aim to select a subset of a larger labeled training set, where the goal is to reduce the amount of training time needed~\cite{loshchilov2015online,katharopoulos2018not,mindermann2022prioritized}. 

In the context of supervised transfer learning, \cite{azizpour2015factors} demonstrated that transfer performance is dependent on the similarity between the source and target domains.  
\cite{Cui_2018_CVPR} proposed a method for selecting a subset of labeled images from a large source collection such that they would be a useful pre-training source for a model that is fine-tuned on a smaller target dataset of interest. 
Their approach estimates domain similarity between class prototypes from the source and target datasets using the Earth Mover's Distance and then greedily selects images from the categories in the source that are most similar to the target.  
There are two issues associated with applying their method to the discovery problem. 
First, we do not have labels for the images in the unlabeled set, and as we will demonstrate later, selecting the most similar categories between the source and the target actually hurts discovery performance. 

Also partially related to our work are methods that have explored the use of different granularities of supervision during training. 
\cite{cole2022label} showed that the precise label hierarchy used during supervised pre-training can have a large impact on object localization performance. 
Specifically, they showed in the context of hierarchically labeled data that using labels that are too coarse (\eg using a generic label such as  `bird') or too fine (\eg using a specific bird species as the label) performs worse than intermediate supervision. 
The impact of label granularity has also been explored in tasks such as action~\cite{shao2020finegym} and attribute-based recognition~\cite{guo2019imaterialist}. 
Works that study the transferability of pre-trained models also exist~\cite{bao2019information,tran2019transferability,nguyen2020leep,bolya2021scalable}, but they cannot be used for category discovery as they are aiming to select models, not data, from a large model zoo.
Finally, some works have shown that features from networks trained using self-supervision on ImageNet~\cite{russakovsky2015imagenet} are effective for discovery~\cite{zhao2021novel,zhao2023incremental}. 
However, to the best of our knowledge, the impact of choosing different labeled data sources for category discovery has not been investigated.

\section{Method}
\label{sec:method}
\vspace{-5pt}
\subsection{Problem statement}
\vspace{-5pt}
In generalized category discovery we are provided with a source labeled dataset $\mathcal{D}^l = \{(\mathbf{x}_i, y_i)\}_{i=1}^{N} \in \mathcal{X}\times \mathcal{Y}_l$ and a target unlabeled dataset $\mathcal{D}^u = \{(\mathbf{x}_j, y_j)\}_{j=1}^{M}\in \mathcal{X}\times\mathcal{Y}_u$. 
In our case, $\mathbf{x}_i$ is an image and $y_i$ is its corresponding discrete category label. 
The labels $y_j$ for the unlabeled dataset are \emph{not} available during training.  
The images in the unlabeled dataset can contain  novel categories that are not present in the labeled set, $\mathcal{Y}_u\setminus\mathcal{Y}_l\neq\emptyset$. 
We also do not know the number of possible categories present in the unlabeled set $|\mathcal{Y}_u|$. 
Importantly, in the generalized setting~\cite{vaze2022generalized}, there can also be images in the unlabeled set from categories that are also present in the labeled set, $\mathcal{Y}_l\subset\mathcal{Y}_u$. 
The task is to develop a feature extractor $f(.)$ that, when combined with a classifier $g(.)$, correctly assigns category labels to the instances in the unlabeled dataset, \ie $\mathbf{p}_i = g(f(\mathbf{x}_i))$, where $\mathbf{p}_i$ is a distribution over category labels from both the unlabeled and labeled sets. 
A successful discovery method correctly classifies unlabeled images as being either categories from the labeled set, or groups them into novel categories.

A growing number of methods have been proposed to address the discovery problem~\cite{vaze2022generalized,cao22,wen2023parametric,pu2023dynamic}. 
Instead of developing a novel discovery method, we ask an orthogonal question. 
Specifically, we ask: ``\textit{What impact does the choice of the labeled dataset $\mathcal{D}^l$ have on downstream discovery performance?}'' 
The motivation for this question is that in category discovery, not only do we not know the number of novel categories in the unlabeled data in advance~\cite{zhao2023learning}, but we also may not even know what `types' of categories are present in the unlabeled data. 
Specifically, the scenario we are considering in this work is one where we are given a large source collection of labeled data and we have to automatically select a subset of it to supervise a model for category discovery on a specific unlabeled target dataset.
To address this, we propose two new approaches for automatically selecting a labeled training dataset to significantly improves discovery performance, all without altering the underlying discovery algorithm used.

\subsection{Category discovery}
For our main experiments, we make use of the recent SimGCD~\cite{wen2023parametric} discovery approach. 
We chose this method because it is conceptually simple and obtains state-of-the-art performance across a range of challenging datasets. 
We also later demonstrate that our observations hold for other discovery methods (\ie GCD~\cite{vaze2022generalized}, XCon~\cite{fei2022xcon}, and $\mu$GCD~\cite{vaze2023no}). 
In this section, we briefly describe the key components of SimGCD, which consists of two main parts: (i) representation learning and (ii) classifier learning. 

\noindent{\bf Representation learning.} 
To learn a discriminative representation that can separate and discover novel categories, SimGCD employs a pre-trained feature extractor $f(.)$ and fine-tunes it with contrastive learning losses.
Given two random augmentations $\hat{\mathbf{x}}_i$ and $\tilde{\mathbf{x}}_i$ of an image $\mathbf{x}_i$ in a training mini-batch $B$ that contains both labeled and unlabeled data, the \emph{self-supervised} contrastive loss can be computed as: 
\begin{equation}
    \mathcal{L}_{\text{rep}}^{u}=\frac{1}{|B|} \sum_{\mathbf{x}_i\in B} - \log \frac{\exp (\hat{\mathbf{z}}_i^\top \tilde{\mathbf{z}}_i / \tau_u)}{\sum_{\mathbf{x}_j \in B} \exp (\hat{\mathbf{z}}_j^\top \tilde{\mathbf{z}}_i / \tau_u)},
\end{equation}
where $\mathbf{z}_i=m(f(\mathbf{x}_i))$ is a projected feature for contrastive learning, $m(.)$ is the projection head, and $\tau_u$ is a temperature value.
In addition to this self-supervised contrastive loss, SimGCD also makes use of a \emph{supervised} contrastive loss to learn a discriminative representation from the labeled data:
\begin{equation}
    \mathcal{L}_{\text{rep}}^{s} = \frac{1}{|B^l|}\sum_{\mathbf{x}_i\in B^l}\omega_{i}\frac{1}{|\mathcal{N}_i|}\sum_{p\in\mathcal{N}_i}-\log\frac{\exp (\hat{\mathbf{z}}_i^\top \tilde{\mathbf{z}}_p / \tau_s)}{\sum_{n\neq i}\exp (\hat{\mathbf{z}}_i^\top \tilde{\mathbf{z}}_n / \tau_s)},\label{eq:rep_sup}
\end{equation}
where $\mathcal{N}_i$ is the indexes of all images in the mini-batch that have the same label as $\mathbf{x}_i$, $\tau_s$ is an additional temperature value, and $\omega_i$ is a weight factor for the contribution of $\mathbf{x}_i$ to the learning process. 
We set the weight to $1$ by default, and later discuss the impact of changing it to `select' different data subsets.
The representation learning training objective is the combined loss: $\mathcal{L}_{\text{rep}}=(1-\lambda) \mathcal{L}_{\text{rep}}^{u} + \lambda \mathcal{L}_{\text{rep}}^{s}$, where $\lambda$ is a weighting hyperparameter.

\noindent{\bf Classifier learning.}
SimGCD employs a parametric classifier to assign labels to images. 
The classifier is trained in a self-distillation fashion where the number of categories $K = |\mathcal{Y}_u|$ is given \textit{a-prior}, or can be estimated using off-the-shelf methods, and
SimGCD randomly initializes a set of prototypes for each category $\mathcal{C}=\{\mathbf{c}_1, \mathbf{c}_2, \dots, \mathbf{c}_K\}$. 
In the training process, a soft label $\hat{\mathbf{p}}_i$ is  calculated by performing softmax classification using the prototypes:
\begin{equation}
    \hat{\mathbf{p}}_i^k = \frac{\exp\left(\frac{1}{\tau_s}({\hat{\mathbf{h}}_i} / {\|\hat{\mathbf{h}}_i\|_2})^\top ({\mathbf{c}_k}/{\|\mathbf{c}_k\|_2})\right)}{\sum_{j}\exp{\left(\frac{1}{\tau_s} (\hat{\mathbf{h}}_i/\|\hat{\mathbf{h}}_i\|_2)^\top (\mathbf{c}_j/\|\mathbf{c}_j\|_2)\right)}},
\end{equation}
where $\hat{\mathbf{h}}_i=f(\hat{\mathbf{x}}_i)$ is the representation of $\hat{\mathbf{x}}_i$ directly from the backbone $f(.)$. 
In addition, a soft label vector $\tilde{\mathbf{p}}_i$ is also generated  for $\tilde{\mathbf{x}}_i$.
A cross-entropy and self-distillation with entropy-regularization loss are used to train this classifier:
\begin{align}
    \mathcal{L}_{\text{cls}}^l &= \frac{1}{|B^l|}\sum_{\mathbf{x}_i\in B^l} \omega_{i}\mathcal{L}_{\text{ce}}(\mathbf{y}_i, \hat{\mathbf{p}}_i), \label{eq:cls_sup}\\
    \mathcal{L}_{\text{cls}}^u &= \frac{1}{|B|}\sum_{\mathbf{x}_i\in B} \mathcal{L}_{\text{ce}}(\tilde{\mathbf{p}}_i, \hat{\mathbf{p}}_i) - \epsilon H(\overline{\mathbf{p}}),
\end{align}
where $\mathbf{y}_i$ is the ground truth label for the labeled example $\mathbf{x}_i$, $\mathcal{L}_{\text{ce}}$ is the cross-entropy loss, $\omega_i$ is the weight factor defaulted to $1$, and $H(\overline{\mathbf{p}})=-\sum \overline{\mathbf{p}} \log \overline{\mathbf{p}}$ regularizes the mean prediction $\overline{\mathbf{p}}=\frac{1}{2|B|}\sum_{\mathbf{x}_i\in B}(\hat{\mathbf{p}}_i + \tilde{\mathbf{p}}_i)$ in a mini-batch.
The final classifier loss is defined as $\mathcal{L}_{\text{cls}}=(1-\lambda) \mathcal{L}_{\text{cls}}^u+\lambda \mathcal{L}_{\text{cls}}^l$. 

The final overall SimGCD training loss is the combination of the representation learning and classifier learning losses:
\begin{equation}
    \mathcal{L} = \mathcal{L}_{\text{rep}} + \mathcal{L}_{\text{cls}}.
\end{equation}

\vspace{-14pt}
\subsection{Data selection} 
\label{sec:selection}
Our central hypothesis is that not all labeled data is equally useful for performing category discovery. 
To address this, we aim to `select' a subset of the entire labeled source dataset to use for training the discovery model (\eg SimGCD). 
We perform data selection by generating instance-specific weights for each data point in the \emph{labeled} source dataset. %
Specifically, the selection process aims to generate a weight $\omega_i\in [0,+\infty)$ for each data point $\mathbf{x}_i$ in the labeled dataset which is used to weight some components of the training loss in~\cref{eq:cls_sup,eq:rep_sup}. %
Thus, data points with weight $0$ are effectively discarded during training, and non-zero weighted data points influence training based on the magnitude of their weight. 
Similarly, using a weight of~$1$ for all data points is equivalent to running off-the-shelf discovery with no weighting.   
Next, we review a related supervised selection approach proposed for improving transfer learning and then introduce two new unsupervised selection methods.  

\vspace{-14pt}
\subsubsection{Supervised data selection.} 
In the context of supervised classification, \cite{Cui_2018_CVPR} proposed a method for dataset selection where the goal was to improve transfer learning performance. 
Given a large labeled source dataset $\mathcal{D}^s$ and a \emph{labeled} target dataset $\mathcal{D}^t$, their method automatically selects a subset of the source data for pre-training an image classifier. 
The intuition is that if they can identify a subset of the source data that is most similar to the target dataset, it will provide a better pre-training signal for a model that can then be fine-tuned on the target dataset. 
They propose a method that uses the Earth Mover's Distance (EMD)  to measure the similarity between the source and target. 
As the number of images in the source dataset could be very large, they start with a simplifying assumption that each category can be represented by a single per-category mean feature vector, which we denote as:
\begin{equation}
\bar{\mathbf{h}}_c = \frac{1}{N_c} \sum_{\substack{(\mathbf{x}, y) \in \mathcal{D} \\ y=c}}f(\mathbf{x}),  
\label{eqn:sup_cluster}
\end{equation}
where $N_c$ is the number of instances of category $c$ in a dataset $\mathcal{D}$. 
This mean category vector computation is only possible because both the source and target datasets are fully labeled. 
In the more challenging category discovery setting, which we will discuss later, the target data is unlabeled. 

Given the per-category mean feature vectors they compute the EMD between two datasets as: 
\begin{equation}
\text{EMD}(\mathcal{D}^s, \mathcal{D}^t) = \frac{\sum_{i=1}^N\sum_{j=1}^M k_{i,j} d_{i,j}}{\sum_{i=1}^N\sum_{j=1}^M k_{i,j}},
\end{equation}
where the distance between pairs of mean vectors is $d_{i,j} =  \lVert \bar{\mathbf{h}}^s_i - \bar{\mathbf{h}}^t_j \rVert_2$, $k_{i,j}$ is the optimal flow that minimizes the overall cost, and $N$ and $M$ are the total number of categories in the source and target datasets. 
From this, they define a `domain similarity' score between two datasets which is represented as:
\begin{equation}
\text{SIM}(\mathcal{D}^s, \mathcal{D}^t) = \exp\left({-\gamma \text{EMD}(\mathcal{D}^s, \mathcal{D}^t)}\right),
\label{eqn:domain_sim}
\end{equation}
where $\gamma$ is a scaling hyperparameter. 

To select a subset of the source dataset for pre-training they perform a greedy selection process whereby they incrementally select the categories from the source that are most similar to the target. 
As this comparison is performed on the per-category mean feature vectors, to construct the final updated source dataset they select each of the images from the source that belong to the selected categories.  
Their approach can be viewed as assigning a weight of in $\{0,1\}$ to each data point in the labeled dataset based on the category it belongs to.

\begin{figure*}[t]
\centering
\includegraphics[width=\textwidth]{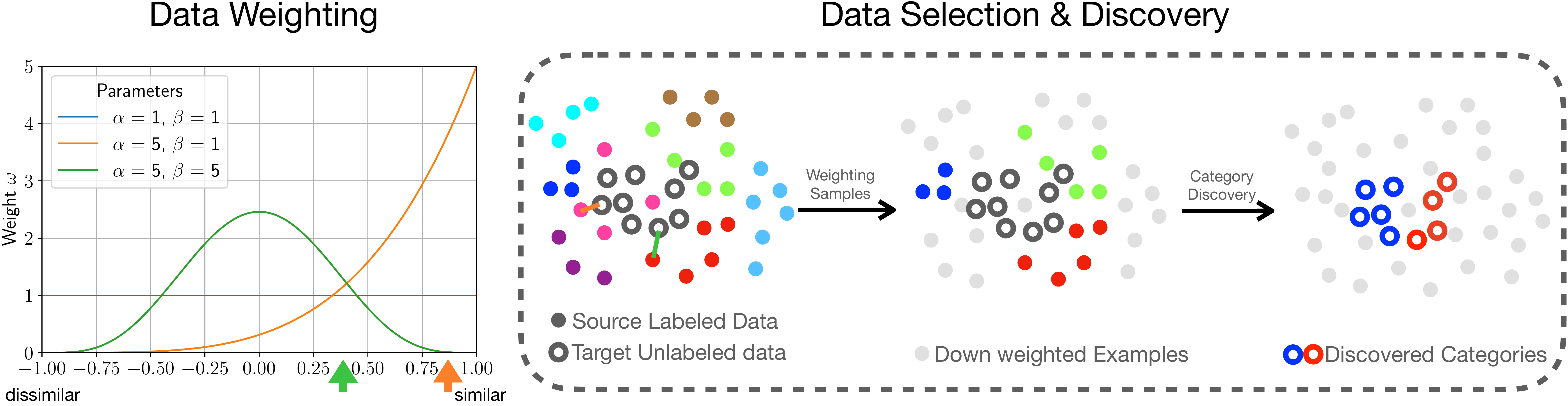}
\vspace{-15pt}
\caption{
(Left) The beta distribution is used to generate weights for the labeled data.
(Right) Overview of our labeled data selection process for category discovery. 
The labeled data is weighted based on the distance to the unlabeled data. The weight is used to change the influence of labeled categories during training. 
The {\color{mygreen}green} and {\color{orangearrow}orange} arrows on the left panel denote the distance between labeled to unlabeled data for two instances illustrated using same color in the first step on the right panel.
}
\label{fig:method_fig}
\vspace{-16pt}
\end{figure*}

\vspace{-14pt}
\subsubsection{Unsupervised data selection.} 
In later category discovery experiments (\cref{tab:diff_source_res}) we demonstrate that the most informative labeled data to use is neither the most similar nor the most dissimilar to the \emph{unlabeled} target.
Instead, for discovery, it is more effective to select labeled data that shares high-level features with the target data such that learned features are helpful for discovery, and at the same time not too similar to cause confusion. 
If they were too similar, the model would likely incorrectly classify the novel categories in the unlabeled target as previously seen categories from the labeled dataset.

There are two main issues associated with applying the previously described selection approach from~\cite{Cui_2018_CVPR} to the problem of category discovery: 
(i) we do not have labels for the images in the unlabeled target data and (ii) the most similar categories from the source are not the most useful for category discovery (see~\cref{fig:intro_fig}).  
Next, we will outline two new conceptually simple unsupervised selection methods that address both of these challenges, where selection is performed by assigning generated weights (either hard or soft) to the labeled data.

\noindent{\bf Binning.} 
One approach for data selection is to simply `bin' the labeled dataset into several equal-sized subsets based on the similarity to the unlabeled target, and then select data from the bins that are not too similar or too dissimilar to the target. 
However, the difficulty is that without labels we cannot use the approach from~\cite{Cui_2018_CVPR} to determine similarity, thus we cannot determine how many bins to use or which bins to discard. 
To address this, we design an unsupervised selection method that first filters out unrelated data and then performs selection. 

We first perform $k$-means clustering~\cite{macqueen1967some_kmeans} on the target unlabeled dataset to obtain an initial clustering, and then compute a per-cluster mean feature vector:
\begin{equation}
    \bar{\mathbf{h}}_c^u = \frac{1}{N_c^u} \sum_{\substack{(\mathbf{x}, \hat{y})\in \mathcal{D}^u \\ \hat{y}=c}}f(\mathbf{x}),
\end{equation}
where $\hat{y}$ is the clustering label generated by running $k$-means on the unlabeled target dataset $\mathcal{D}^u$, and $N_c^u$ is the number of images in cluster $c$.
As in~\cref{eqn:sup_cluster}, we can directly use the labels to generate the per-category vectors $\bar{\mathbf{h}}_c^l$ for $\mathcal{D}^l$.

The key for the data selection is to generate lower weights for data that are too dissimilar or too similar.
Our binning approach achieves this by first discarding dissimilar samples and then removing data that is too similar. 
To discard dissimilar samples, we set a similarity threshold to filter out distant unrelated samples. 
The threshold is determined by computing the similarities between the data points \emph{within} the target unlabeled data. 
This gives us a measure of how similar the categories in the target unlabeled data should be, and thus we can filter out data that is less similar to the target data than this threshold.
Next, we need to remove data that is too \emph{similar}. 
To do this, we rank the labeled categories in the remaining source data based on their similarity to the target data and then select the labeled categories that are the most dissimilar from the target unlabeled data. 
This operation removes data that is too similar to the target data.
The above selection is performed by setting the weight factor to a `hard' value, \ie $\omega \in \{0,1\}$.
In practice, we can discard the samples with weight $\omega_i=0$ during training as they do not directly impact the loss. 
A more detailed description is provided in the supplementary material.

\noindent{\bf Soft weighting.} The issue with the previous approach is that the choice of `hard' binning enforces a hard selection, whereby some labeled data is kept and the rest is discarded. 
Additionally, the binning method requires performing clustering on the target unlabelled data first, which requires knowledge (or an estimate) of how many clusters there should be in the unlabelled target data. 
An alternative approach is to use soft weights to downweight data that is either too similar or too dissimilar.
This intuition motivates us to introduce a new approach for soft selection that leverages the functional form of the beta distribution. 
Specifically, the probability density function of the beta distribution is defined as: 
\begin{equation}
    b(x;\alpha,\beta) = \frac{1}{B(\alpha, \beta)}x^{\alpha-1}(1-x)^{\beta-1},
\end{equation}
where $B(\alpha, \beta)$ is the beta function which ensures the distribution sums to $1$, $\alpha,\beta >0$ are shape parameters, and $0\leq x\leq 1$ is the input similarity.

We can use this distribution as a mechanism to generate weights. 
Specifically, we can input a similarity score in the range $[-1, 1]$ to $b(.;\alpha,\beta)$, and use the output as the weight for the respective data point from the labeled source. 
Using different shape parameters with for the distribution changes which data is weighted higher or lower. 
We compute the similarity score as the cosine similarity of each of labeled centroids to the farthest unlabeled data point:
\begin{equation}
    \gamma_c^l = \min_{\mathbf{x}\in \mathcal{D}^u} \cos(\bar{\mathbf{h}}_c^l, f(\mathbf{x})),
\end{equation}
where $\cos(.,.)$ denotes the cosine similarity function.
The intuition for choosing the farthest unlabeled data point is to see how related the labeled category is to the farthest unlabeled data, an ablation is provided in the supplementary for this design choice.
Compared to the selection used in~\cite{Cui_2018_CVPR} and our hard binning method, our soft weighting approach is more lightweight as it does not require running a clustering first on the unlabeled data.
The weight of the labeled data is thus calculated as:
\begin{equation}
    \omega^c = b(\gamma_c^l;\alpha, \beta),\label{eq:beta_weight}
\end{equation}
where $\omega^c$ is a weight value that is assigned to all data points from the category $c$, \ie $\omega_i = \omega^c, \forall (\mathbf{x}_i, y)\in \mathcal{D}^l \text{ and } y=c$.
This formulation gives us a simple way of generating smaller weights for data points that are too similar or dissimilar.
Additionally, this formulation allows us to further tune the shape of the curve by changing the shape parameters $\alpha$ and $\beta$, as shown in~\cref{fig:method_fig} left.
For example, setting both $\alpha$ and $\beta$ to $1$ yields a degenerated solution of uniformed weight (as in conventional discovery).
When $\alpha, \beta > 1$ and $\alpha>\beta$ the algorithm will favor more similar data points to the target (as in~\cite{Cui_2018_CVPR}),  when $\alpha, \beta > 1$ and $\beta > \alpha$ it will favor more dissimilar ones, and when $\alpha, \beta > 1$ and $\beta = \alpha$ it will favor data points that are not too similar or not too dissimilar.

\vspace{-12pt}
\section{Experiments}
\label{sec:exps}
\vspace{-10pt}
Here, we quantify the impact of labeled data on category discovery performance and present the experimental results of our proposed data selection approaches.

\vspace{-12pt}
\subsection{Implementation details}
\vspace{-8pt}
For the majority of our experiments, we use the recent state-of-the-art generalized category discovery method SimGCD~\cite{wen2023parametric}. 
We use the DINOv2~\cite{oquab2023dinov2} pre-trained ViT-Base/14 model~\cite{dosovitskiy2020image} for our backbone~$f(.)$. 
Following the standard practice in category discovery, we freeze the backbone model up to the last block to avoid overfitting~\cite{vaze2022generalized,wen2023parametric}.
As a baseline, we also report the discovery performance resulting from running `$k$-means' clustering on the DINOv2 features using the known number of categories for $k$.
We evaluate on the commonly used Semantic-Shifts Benchmarks (SSB)~\cite{vaze2021open}. 
SSB contains data from CUB~\cite{cub200}, Stanford-Cars~\cite{stanfordcars}, and FGVC-Aircraft~\cite{aircraft}, and provides the category splits of the data for category discovery with an annotation of `Easy', `Medium', and `Hard' categories that are built on the hierarchy from the dataset labels. The `Easy' categories are those that can be easily separated from the labeled categories, similar for `Medium' and `Hard' categories. 
In this paper, we denote `Hard' categories as `Similar' and `Easy' categories as `Dissimilar'. 
We use these annotations in the evaluation and motivation of the data selection problem. 
Additionally, we also leverage the Stanford-Dogs~\cite{stanforddogs} dataset and a subset of Insect categories from iNat2021~\cite{inat21}.
The Stanford-Dogs dataset is also split into `Similar', `Medium', and `Dissimilar' categories using its label hierarchy.  

\noindent{\bf Evaluation protocol.}
To evaluate model performance, we adopt the commonly used clustering accuracy metric~\cite{vaze2022generalized}.
During the evaluation, given the ground truth labels $y$ and the model predicted labels $\tilde{y}$, we calculate the accuracy as $\text{ACC}=\frac{1}{M}\sum_{i=1}^M \mathds{1}(y^*_i = \mathcal{P}(\tilde{y}_i))$, where $M= |\mathcal{D}^u|$, and $\mathcal{P}$ is the optimal permutation that matches the predicted cluster assignments to the ground truth class labels. 
Additionally, we report `All', `New', and `Old' performance as is common in the literature.
`All' is for the performance on all clusters, and `New' and `Old' denote the performance on novel and seen categories respectively.
We report the average of three different runs in each table.

\vspace{-10pt}
\subsection{Evaluation}

\begin{wraptable}{L}{.5\linewidth}
    \vspace{-1.1cm}
    \caption{Performance on `New' categories when using different labeled data subsets to supervise SimGCD~\cite{wen2023parametric}. 
    }
    \label{tab:diff_source_res}
    \resizebox{\linewidth}{!}{
    \begin{tabular}{lcccc}
    \toprule
    Labeled data               & CUB  & SCars  & Aircraft  & SDogs \\
    \midrule
    $k$-means                  &  32.1    &  13.8       &    16.8      &   30.2     \\
    \midrule
    Orig. data                   & 57.7 & 45.0    &  51.8    &  58.2   \\
    \midrule
    OOD iNat-Insect            & 3.4  & 3.1     &  4.5     &  2.1          \\
    Similar                    & 47.6 & \textbf{43.4}    &  40.0    &  45.6         \\
    Medium                     & {\bf 79.7} &  \underline{43.2}    &  {\bf  56.6}    &  \textbf{53.1}         \\
    Dissimilar                 & \underline{61.4} & 38.4    &  \underline{47.6}    &  \underline{50.2}         \\
    \bottomrule
    \end{tabular}
    }
    \vspace{-1cm}
\end{wraptable}

In this section, we demonstrate our motivating experiments and report the impact of labeled data selection on category discovery performance.
Unless noted otherwise, the scores are the performance on `New' categories.

\noindent{\bf Labeled data selection.} In~\cref{tab:diff_source_res}, we report the discovery performance of SimGCD~\cite{wen2023parametric} on each of the target datasets when using different source labeled data. 
Note that these `Similar', `Medium', and `Dissimilar' splits are based on the ground truth hierarchical labels of the dataset, thus the results are only for demonstration purposes as in the real world, we cannot obtain this ground truth information. 
We observe that the choice of labeled data has a large impact on discovery performance. 

\begin{wrapfigure}{R}{0.5\linewidth}
    \vspace{-.6cm}
    \includegraphics[width=\linewidth]{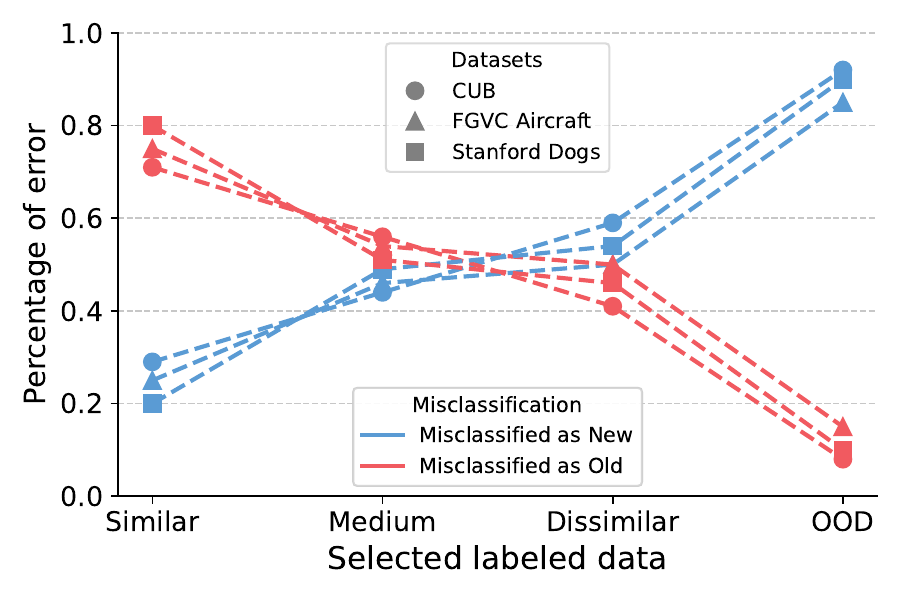}
    \vspace{-.8cm}
    \captionof{figure}{Percentage of errors on three datasets when using `Similar', `Medium', `Dissimilar', and `OOD' labeled training sets. 
    `Misclassified as New' means that an example from a `New' category is assigned to another `New' one, and `Misclassified as Old' means it is assigned to an `Old' one. %
    }
    \vspace{-.8cm}
    \label{fig:error_source}
\end{wrapfigure}

Perhaps surprisingly, the source split that performs the best (`Medium') is a subset of the source data that contains categories that are not too similar nor too dissimilar.  
Using the `Medium' labeled data alone can match, or even outperform,  using all labeled data in some cases (\ie on CUB and Aircraft), despite using only one-third of the data. 
By taking a closer look at the source of errors in~\cref{fig:error_source}, we identify that when the model is using the `Similar' splits, the majority of errors stem from misclassifying the novel categories to previously seen ones. 
When the model is trained with `Dissimilar' splits, the error is mainly due to misclassifying the novel categories to incorrect novel clusters.
When the labeled split comes from Out-of-Domain (OOD) insects, the error is almost entirely from misclassifying novel categories as incorrect novel clusters. 
These results indicate that when the model is trained with `Dissimilar' or OOD data, it can learn to predict the novel categories as new clusters, but cannot learn a discriminative enough representation to separate the novel categories apart. 
On the other hand, when the model is trained with `Similar' data, it struggles to separate novel categories from seen ones, resulting in a higher error when misclassifying novel categories as seen.

\begin{table*}[t]
\centering
\caption{Category discovery results on the Semantic Shift Benchmark~\cite{vaze2021open}. Here, labeled data is selected from {\bf all} the datasets combined, \ie we combined all the labeled data from the four datasets and the OOD insects. Performance change with respect to no-selection in presented in parentheses.
} 
\vspace{-8pt}
\resizebox{\linewidth}{!} 
{
\begin{tabular}{l lll lll lll lll}
\toprule
&   \multicolumn{3}{c}{CUB} & \multicolumn{3}{c}{Stanford Cars} & \multicolumn{3}{c}{FGVC-Aircraft} & \multicolumn{3}{c}{Stanford Dogs}\\
\cmidrule(rl){2-4}\cmidrule(rl){5-7}\cmidrule(rl){8-10}\cmidrule(rl){11-13}
Methods        & All  & Old  & New  & All  & Old  & New  & All  & Old  & New  & All &  Old & New\\
\midrule
$k$-means~\cite{macqueen1967some_kmeans}  
& 34.7 & 49.8 & 34.2 & 16.7 & 14.5 & 17.1 & 17.9 & 16.5 & 19.2 & 21.5 & 22.6 & 20.1\\
\midrule
GCD~\cite{vaze2022generalized}  
& {44.8}                 & {50.9}                 & {41.2}             & {35.9}             & 55.2             & {26.7}    & 
                        {41.3}             & {38.7}             & {45.8}             & 48.7               & 54.1 & 46.1\\
GCD w/ \cite{Cui_2018_CVPR} 
&  40.7\perfdown{-4.1}   & 46.8\perfdown{-4.1}    & 36.8\perfdown{-4.4}& 35.0\perfdown{-0.9}& 55.9\perfup{+0.7}& 25.7\perfdown{-1.0}  &  34.8\perfdown{-6.5}& 30.1\perfdown{-8.6}& 36.9\perfdown{-8.9}& 44.7\perfdown{-4.0} & 50.6\perfdown{-3.5} & 42.7\perfdown{-3.4}\\
GCD w/ Bins
&  47.9\perfup{+3.1}   &   52.8\perfup{+1.9}    &    44.2\perfup{+3.0}   &  36.0\perfup{+0.1}    &  55.0\perfdown{-0.2}  & 25.0\perfdown{-1.7} &  44.9\perfup{+3.6}    &  40.8\perfup{+2.1}   &   47.2\perfup{+1.4} & 49.4\perfup{+0.7} & 54.0\perfdown{-0.1} & 48.2\perfup{+2.1} \\
GCD w/ Beta
&  47.6\perfup{+2.8}   &   53.0\perfup{+2.1}    &    43.7\perfup{+2.5}   &  34.9\perfdown{-1.0}    &  56.2\perfup{+1.0}  & 25.5\perfdown{-1.2} &  44.1\perfup{+2.8}    &  39.7\perfup{+1.0}   &   46.2\perfup{+0.4} & 49.9\perfup{+1.2} & 55.5\perfup{+1.4} & 47.4\perfup{+1.3} \\
\midrule
XCon~\cite{fei2022xcon}          
&  46.7  &  50.4  & 44.3   &  38.7  & 55.9   &  28.7  &  43.1  &  40.7  &  47.9  & 50.0 & 54.4 &  48.0   \\
XCon w/ \cite{Cui_2018_CVPR} 
&  42.8\perfdown{-3.9}   & 45.7\perfdown{-4.7}  & 39.6\perfdown{-4.7}   &  39.2\perfup{+0.5}  & 55.0\perfdown{-0.9}   &  26.9\perfdown{-1.8}  &  35.8\perfdown{-7.3}  &  31.0\perfdown{-9.7}  &  40.8\perfdown{-7.1}  & 45.9\perfdown{-4.1} & 52.0\perfdown{-2.4} & 42.2\perfdown{-5.8}\\
XCon w/ Bins
&  48.2\perfup{+1.5}   &  51.0\perfup{+0.6} & 46.7\perfup{+2.4}   &  39.8\perfup{+1.1}  & 55.7\perfdown{-0.2}   &  27.9\perfdown{-0.8}  &  44.8\perfup{+1.7}  &  42.1\perfup{+1.4}  &  48.3\perfup{+0.4}  & 52.3\perfup{+2.3} & 56.1\perfup{+1.7} & 48.9\perfup{+0.9} \\
XCon w/ Beta
&  47.8\perfup{+1.1}   &  51.8\perfup{+1.4} & 46.0\perfup{+1.7}   &  39.5\perfup{+0.8}  & 56.6\perfup{+0.7}   &  28.9\perfup{+0.2}  &  45.0\perfup{+1.9}  &  42.6\perfup{+1.9}  &  49.3\perfup{+1.4}  & 52.5\perfup{+2.5} & 56.3\perfup{+1.9} & 49.4\perfup{+1.4} \\
\midrule
SimGCD~\cite{wen2023parametric} 
& {55.1}               & {61.2}                & {52.3}               & {50.4}           &{68.3}            &{40.9}          & {51.8}                  & {56.2}            &{49.3}             &  57.2  &  60.1 &  54.2 \\
SimGCD w/ \cite{Cui_2018_CVPR} 
&  50.3\perfdown{-4.8} & 58.3\perfdown{-2.9}   &  46.5\perfdown{-5.8} &  51.6\perfup{+1.2}&69.4\perfup{+1.1}&44.7\perfup{+3.8}  &  44.2\perfdown{-7.6}  &48.7\perfdown{-7.5}&40.7\perfdown{-8.6} & 55.2\perfdown{-2.0}  & 57.2\perfdown{-2.9}  & 50.1\perfdown{-4.1} \\
SimGCD w/ Bins
& 58.2\perfup{+3.1}   & 64.6\perfup{+3.4}    &  56.7\perfup{+4.4}    &  52.4\perfup{+2.0} & 68.2\perfdown{-0.1} &45.9\perfup{+5.0} & 55.7\perfup{+3.9}  & 57.9\perfup{+1.7}    &   56.8\perfup{+7.5}  & 58.1\perfup{+0.9}   &  62.3\perfup{+2.2}        &   55.6\perfup{+1.4} \\
SimGCD w/ Beta
& 58.6\perfup{+3.5}   & 64.5\perfup{+3.3}    &  56.8\perfup{+4.5}    &  52.8\perfup{+2.4} & 68.9\perfup{+0.6} &46.3\perfup{+5.4} & 56.3\perfup{+4.5}  & 58.3\perfup{+2.1}    &   56.9\perfup{+7.6}  & 58.4\perfup{+1.2}   &  62.8\perfup{+2.7}        &   56.6\perfup{+2.4} \\
\midrule
$\mu$GCD~\cite{vaze2023no} 
& {59.3}               & {67.8}                & {58.4}               & {58.6}           &{72.3}            &{48.3}          & {55.9}                  & {63.5}            &{54.2}             &  63.4  &  65.1 &  58.3 \\
$\mu$GCD w/ \cite{Cui_2018_CVPR} 
&  54.3\perfdown{-5.0} & 62.4\perfdown{-5.4}   &  53.5\perfdown{-4.9} &  54.6\perfdown{-4.0}&68.1\perfdown{-4.2}&45.1\perfdown{-3.2}  &  49.1\perfdown{-6.8}  &56.9\perfdown{-6.6}&48.7\perfdown{-7.5} & 56.2\perfdown{-7.2}  & 59.8\perfdown{-5.3}  & 54.1\perfdown{-4.2} \\
$\mu$GCD w/ Bins   
& 61.2\perfup{+1.9}& 69.3\perfup{+1.5} & 60.1\perfup{+1.7} & 59.5\perfup{+0.9}& 73.5\perfup{+1.2}& 50.4\perfup{+2.1}& 57.2\perfup{+1.3}& 64.5\perfup{+1.0}& 55.4\perfup{+1.2}& 64.5\perfup{+1.1}& 66.3\perfup{+1.2}& 59.6\perfup{+1.3} \\
$\mu$GCD w/ Beta
& 62.3\perfup{+3.0}   & 70.5\perfup{+2.7}    &  61.8\perfup{+3.4}    &  60.4\perfup{+1.8} & 74.3\perfup{+2.0} &51.0\perfup{+2.7} & 58.2\perfup{+2.3}  & 65.8\perfup{+2.3}    &   57.7\perfup{+3.5}  & 66.4\perfup{+3.0}   &  67.8\perfup{+2.7}        &   61.3\perfup{+3.0} \\
\bottomrule
\end{tabular}
}
\label{tab:ssb_all}
\vspace{-8pt}
\end{table*}

\begin{table}[t]
    \centering
    \caption{Discovery results on  NABirds~\cite{van2015building}, Herbarium-19~\cite{tan2019herbarium}, and ImageNet~\cite{russakovsky2015imagenet}. 
    The source data consists of these three datasets plus the source data  used in \cref{tab:ssb_ori}.}
    \vspace{-8pt}
    \label{tab:nabirds_herb19}
    \resizebox{\linewidth}{!} 
    {
    \begin{tabular}{l lll lll lll lll}
    \toprule
    Method     & \multicolumn{3}{c}{NABirds}  & \multicolumn{3}{c}{Herb-19} & \multicolumn{3}{c}{ImageNet-100}  & \multicolumn{3}{c}{ImageNet-1k-SSB}\\
    \cmidrule(lr){2-4}\cmidrule(lr){5-7}\cmidrule(lr){8-10}\cmidrule(lr){11-13}
               &  All & Old & New  & All & Old & New &  All & Old & New  & All & Old & New\\
    \midrule
    SimGCD     &  54.6 & 61.2  & 49.0  & 44.2 & 57.6 & 37.0 & 78.5  &  86.7 &  {\bf 73.5}   & 50.2 &   73.6 &  42.6\\
    SimGCD w/~\cite{Cui_2018_CVPR} & 50.1\perfdown{-4.5} & 54.6\perfdown{-6.6} & 44.8\perfdown{-4.2}  & 38.1\perfdown{-4.1}  & 52.6\perfdown{-5.0} & {\bf 38.7}\perfup{+1.7} & 76.4\perfdown{-2.1}  &  84.2\perfdown{-2.5} &  70.2\perfdown{-3.3}   & 48.2\perfdown{-2.0} &  68.9\perfdown{-4.7} &  40.2\perfdown{-2.4}\\
    SimGCD w/ Bins & {\bf 56.2}\perfup{+1.4} & {\bf 63.5}\perfup{+2.3}  & {\bf 51.6}\perfup{+2.6} & 44.7\perfup{+0.5} & {\bf 58.1}\perfup{+0.5}  & 36.8\perfdown{-0.2} & 78.2\perfdown{-0.3}  &  {\bf 87.3}\perfup{+0.6} &  73.3\perfdown{-0.2}   & 51.4\perfup{+1.2} &  73.5\perfdown{-0.1} &  {43.7}\perfup{+1.1} \\
    SimGCD w/ Beta & 55.7\perfup{+1.1}   &   61.8\perfup{+1.6}  &  50.8\perfup{+1.8}   & {\bf 45.2}\perfup{+1.0} & 57.8\perfup{+0.2} & 38.3\perfup{+1.3}& {\bf 78.9}\perfup{+0.4}  & 87.0\perfup{+0.3} & 73.4\perfdown{-0.1}  &  {\bf 51.7}\perfup{+1.5} & {\bf 74.0}\perfup{+0.4} & {\bf 44.0}\perfup{+1.4} \\
    \bottomrule
    \end{tabular}
    }
    \vspace{-14pt}
\end{table}

\begin{table*}[t]
\centering
\caption{Category discovery results on the Semantic Shift Benchmark~\cite{vaze2021open}. Here, labeled data for a given dataset is selected from the {\bf original} labeled splits for that dataset, \eg we only use labeled data from CUB for the experiments on CUB. 
} 
\vspace{-8pt}
\resizebox{\linewidth}{!} 
{
\begin{tabular}{l lll lll lll lll}
\toprule
&   \multicolumn{3}{c}{CUB} & \multicolumn{3}{c}{Stanford Cars} & \multicolumn{3}{c}{FGVC-Aircraft}  & \multicolumn{3}{c}{Stanford Dogs}\\
\cmidrule(rl){2-4}\cmidrule(rl){5-7}\cmidrule(rl){8-10} \cmidrule(rl){11-13}
Methods        & All  & Old  & New  & All  & Old  & New  & All  & Old  & New  & All & Old & New\\
\midrule
$k$-means~\cite{macqueen1967some_kmeans}  
& 34.3 & 38.9 & 32.1 & 12.8 & 10.6 & 13.8 & 16.0 & 14.4 & 16.8 & 18.9 & 19.1 & 18.0 \\
\midrule
GCD~\cite{vaze2022generalized}       
& {51.3} & {56.6} & {48.7} & {39.0} & 57.6 & {29.9} & {45.0} & 41.1 & {46.9} & 54.6 & 58.7 & 50.1\\
GCD w/ \cite{Cui_2018_CVPR} 
&  45.7\perfdown{-5.6} &   51.2\perfdown{-5.4}  &  44.7\perfdown{-4.0}  &  37.2\perfdown{-1.8} &  56.0\perfdown{-1.6} &  23.7\perfdown{-6.2}  &  40.1\perfdown{-4.9}  & 38.7\perfdown{-2.4} & 40.0\perfdown{-6.9} &  48.9\perfdown{-5.7}  &  50.2\perfdown{-8.5}  & 45.7\perfdown{-4.4} \\
GCD w/ Bins
&  50.2\perfdown{-1.1} &   54.7\perfdown{-1.9} &  49.7\perfup{+1.0}   &  40.3\perfup{+1.3} &  57.7\perfup{+0.1} &  32.0\perfup{+2.1}  &  39.7\perfdown{-5.3}  & 39.9\perfdown{-1.2} & 41.3\perfdown{-5.6} &  53.8\perfdown{-0.8} & 54.3\perfdown{-4.4}  & 50.1\perfup{+0.0} \\
GCD w/ Beta
& 52.3\perfup{+1.0} & 57.8\perfup{+1.2} & 49.5\perfup{+0.8} & 41.2\perfup{+2.2} & 59.0\perfup{+1.4} & 33.4\perfup{+3.5} & 44.7\perfdown{-0.3} & 41.3\perfup{+0.2} & 46.4\perfdown{-0.5} & 55.4\perfup{+0.8} & 59.5\perfup{+0.8} & 50.6\perfup{+0.5}\\
\midrule
XCon~\cite{fei2022xcon}           
&    52.1   &   54.3  & 51.0  &  40.5  & 58.8 & 31.7   & 47.7   & 44.4 & 49.4  & 56.1 & 59.1 & 52.4   \\
XCon w/ \cite{Cui_2018_CVPR} 
&  48.3\perfdown{-3.8}   &   50.2\perfdown{-4.1}  & 47.8\perfdown{-3.2}  &  39.8\perfdown{-0.7} &  57.2\perfdown{-1.6} & 30.2\perfdown{-1.5}   & 43.5\perfdown{-4.2}   & 40.3\perfdown{-4.1}  & 45.2\perfdown{-4.2} &  54.1\perfdown{-2.0}  & 56.7\perfdown{-2.4} & 49.7\perfdown{-2.7} \\
XCon w/ Bins
&  51.7\perfdown{-0.4}   &   52.8\perfdown{-1.5}   &  53.7\perfup{+2.7}  &  40.8\perfup{+0.3}  & 57.1\perfdown{-1..7} & 34.9\perfup{+3.2}   & 48.3\perfup{+0.6}   & 44.0\perfdown{-0.4}  & 49.7\perfup{+0.3} & 58.9\perfup{+2.8}  &  61.2\perfup{+2.1}  & 54.6\perfup{+2.2} \\
XCon w/ Beta
&  52.8\perfup{+0.7}   &   55.5\perfup{+1.2}   &  51.4\perfup{+0.4}  &  41.7\perfup{+1.2}  & 59.9\perfup{+1.1} & 35.9\perfup{+4.2}   & 49.4\perfup{+1.7}   & 46.1\perfup{+1.7}  & 51.3\perfup{+1.9} & 60.0\perfup{+3.9}  &  62.3\perfup{+3.2}  & 55.0\perfup{+2.6} \\
\midrule
SimGCD~\cite{wen2023parametric}                             
& {60.3} & {65.6} & {57.7} & {53.8} & {71.9} & {45.0} & {54.2} & {59.1} & {51.8} &  {62.8}   &  {64.7}     &  {58.2}  \\
SimGCD w/ \cite{Cui_2018_CVPR} 
&  58.9\perfdown{-1.4}&63.6\perfdown{-2.0}&55.4\perfdown{-2.3}&54.9\perfup{+1.1}&72.8\perfup{+0.9}&46.7\perfup{+1.7}&48.9\perfdown{-5.3}&56.3\perfdown{-1.8}&44.3\perfdown{-7.5} & 60.2\perfdown{-2.6} & 65.1\perfup{+0.4} & 50.9\perfdown{-7.3} \\
SimGCD w/ Bins
&63.2\perfup{+3.9}&66.7\perfup{+1.1}&62.8\perfup{+5.1}&53.0\perfdown{-0.8}&68.7\perfdown{-3.2}&44.5\perfdown{-0.5}&59.8\perfup{+5.6}&63.5\perfup{+4.4}&56.0\perfup{+4.2}&64.7\perfup{+1.9} & 66.0\perfup{+1.3} & 58.8\perfup{+0.6}\\
SimGCD w/ Beta
&64.5\perfup{+5.2}&67.9\perfup{+2.2}&64.0\perfup{+6.3}&54.1\perfup{+0.3}&70.1\perfdown{-1.8}&46.3\perfup{+1.8}&60.7\perfup{+6.5}&64.8\perfup{+5.7}&57.1\perfup{+5.3}&66.0\perfup{+3.2} & 67.1\perfup{+2.4} & 59.6\perfup{+1.4}\\
\midrule
$\mu$GCD~\cite{vaze2023no}                             
& {65.7} & {68.0} & {64.6} & {58.7} & {76.5} & {50.2} & {58.7} & {64.7} & {56.8} &  {67.9}   &  {69.4}     &  {62.5}  \\
$\mu$GCD w/ \cite{Cui_2018_CVPR} 
&  64.3\perfdown{-1.4}&65.2\perfdown{-2.8}&60.4\perfdown{-4.2}&57.1\perfdown{-1.7}&75.0\perfdown{-1.5}&48.7\perfdown{-1.3}&54.9\perfdown{-3.8}&63.5\perfdown{-1.2}&52.3\perfdown{-4.5} & 64.2\perfdown{-3.7} & 66.1\perfdown{-3.3} & 58.4\perfdown{-4.1} \\
$\mu$GCD w/ Bins
&67.3\perfup{+1.6}&69.0\perfup{+1.0}&65.7\perfup{+1.1}&60.4\perfup{+1.3}&77.8\perfup{+1.3}&51.5\perfup{+1.3}&60.3\perfup{+1.6}&65.9\perfup{+1.1}&58.9\perfup{+2.1}&69.3\perfup{+1.4} & 71.1\perfup{+1.7} & 64.2\perfup{+1.7}\\
$\mu$GCD w/ Beta
&68.4\perfup{+2.7}&70.1\perfup{+2.1}&66.6\perfup{+2.0}&61.0\perfup{+1.9}&78.6\perfup{+2.1}&52.2\perfup{+2.0}&62.0\perfup{+2.3}&67.2\perfup{+2.4}&59.8\perfup{+3.0}&70.3\perfup{+2.4} & 72.0\perfup{+2.6} & 65.4\perfup{+2.9}\\
\bottomrule
\end{tabular}
}
\vspace{-.3cm}
\label{tab:ssb_ori}
\end{table*}

\noindent{\bf Selecting from a large pool of data.}
Previous category discovery works explore the discovery problem on a target unlabeled dataset using a fixed labeled dataset containing the same visual categories. 
Instead, we consider a more realistic scenario where  we need select a suitable labeled dataset from a large repository of  potentially unrelated datasets. 
Only some of the categories in this repository will be relevant to the unlabeled target. 
In our experiments, we set the pool of datasets to be the combination of the labeled data from CUB, Stanford-Cars, Stanford-Dogs, FGVC-Aircraft, and iNat-Insect, and then evaluate discovery performance on a given target dataset.
The results are presented in~\cref{tab:ssb_all}.  

We observe that using the combined labeled data to train discovery models it actually performs worse than just using the original labeled data for a given target task (see~\cref{tab:ssb_ori}), even if the combined labeled data is three-to-four times larger. 
This suggests that for category discovery, more labeled data may not always be helpful. 
We conduct experiments comparing the data selection method of~\cite{Cui_2018_CVPR} and our proposed approaches (`w/ Bins'  and `w/ Beta').
We observe that~\cite{Cui_2018_CVPR}, which selects similar data to the target, is not suitable for category discovery and it results in a performance drop in most cases. 
In contrast, discovery models trained with our data selection methods can match, or even outperform, models trained with the provided labeled data split, indicating their effectiveness. 
Furthermore, in~\cref{tab:nabirds_herb19}, we present results  on larger and more challenging datasets. 
We observe that our methods still yields better performance compared to the vanilla training baseline, and the improvements on these challenging benchmarks are significant above noise.
Interestingly, on the NABirds dataset the binning method outperforms the beta soft weighting. 
We conjecture this to the very fine-grained nature of this dataset, thus all labeled data has roughly same distance to the target unlabeled data, resulting in similar weights for all labeled data.

\noindent{\bf Selecting from the original labeled data.}
In~\cref{tab:ssb_ori}, we present the results of data selection using the originally provided labeled data for each dataset. 
This is the standard setting explored in existing discovery works.  
The most interesting observation here comes from comparing our data selection methods with the standard approach of directly using all data of the provided labeled dataset.
We can see that in almost all cases, using our soft weighting  selection method (`w/ Beta')  outperforms this baseline. %
Owing to our improvement over the already highly performant  $\mu$GCD~\cite{vaze2023no}, our results set a new state-of-the-art on each dataset. %
This result further strengthens our claim that labeled data matters for category discovery, as even the original provided labeled data is not the optimal one for category discovery. 
Thus, more effort should be put into developing new methods for automatically selecting suitable labeled data subsets for target datasets of interest. 
We also observe that~\cite{Cui_2018_CVPR} is not effective at data selection in the category discovery setting. 
In almost all cases, it decreases performance as it is designed to select the most \emph{similar} data to the target. %

\vspace{-10pt}
\subsection{Ablations}
\vspace{-5pt}
In this section, we present ablation experiments on our proposed method with the beta distribution. More ablations can be found in the supplementary.

\noindent{\bf Different of $\alpha,\beta$ values.}
In~\cref{sec:selection}, we showed that by changing the value of $\alpha$ and $\beta$, we can control the shape of the curve for generating data weighs thus interpolating between different strategies.
In \cref{tab:alpha_beta_main} we observe that when setting the values to $\alpha=5$ and $\beta=1$, to favor more similar samples like in~\cite{Cui_2018_CVPR}, the performance drops.
Setting both $\alpha$ and $\beta$ values to larger than $1$ yields consistent performance improvements, and our default choice of $\alpha=\beta=5$ yields consistently strong performance across the three datasets.

\noindent{\bf Hard threshold cutting.}
In~\cref{sec:selection}, we discussed using the weight generated by~\cref{eq:beta_weight} to weigh the influence of each example. 
\cref{tab:threshold_cut_main} shows the ablation of thresholding the generated weight to $\{0,1\}$.
By thresholding the weight $\omega_i$ of an instance $\mathbf{x}_i$ to $0$, we can effectively remove that sample from the training dataset. Thus the training time can be reduced.
These results demonstrate that while using the soft weight performs the best, removing some examples from training not only does not hurt performance but improves it to some extent.

\begin{table}[h]
\vspace{-.4cm}
\centering
\parbox{.5\linewidth}{
    \centering
    \caption{Ablations on different values of $\alpha$ and $\beta$. We default to $\alpha=\beta=5$ for our experiments. Scores are performance on `New' categories.}
    \label{tab:alpha_beta_main}
    \vspace{-.35cm}
    \resizebox{0.45\columnwidth}{!}{
    \begin{tabular}{cc| lll}
    \toprule
    $\alpha$  & $\beta$    &  CUB  &  Aircraft  & SDogs \\
    \midrule
    1 & 1 &   52.3    &   49.3     & 54.2 \\
    5 & 1 &   44.5\perfdown{-7.8}  &   39.5\perfdown{-9.8}      & 48.9\perfdown{-5.3} \\
    5 & 3 &   57.5\perfup{+5.2}    &   56.4\perfup{+7.1}     & 56.2\perfup{+2.0}\\
    3 & 5 &   57.3\perfup{+5.0}    &   56.7\perfup{+7.4}     & 56.7\perfup{+2.5}\\
    5 & 5 &   56.8\perfup{+4.5}    &   56.9\perfup{+7.6}     & 56.6\perfup{+2.4} \\
    \bottomrule
    \end{tabular}
    }
}
\hfill
\parbox{.48\linewidth}{
\vspace{-.3cm}
    \centering
    \caption{%
    Our main experiments use soft weighting, here we ablate hard thresholding those weights. Scores are performance on `New' categories.}
    \label{tab:threshold_cut_main}
    \vspace{-.35cm}
    \resizebox{0.5\columnwidth}{!}{
    \begin{tabular}{l| lll}
    \toprule
      &  CUB  &  Aircraft  & SDogs \\
    \midrule
    baseline      & 52.3    &   49.3     & 54.2 \\
    threshold=0.2 & 55.0\perfup{+2.7} & 55.8\perfup{+6.5} & 56.0\perfup{+1.8} \\
    threshold=0.5 & 54.7\perfup{+2.4} & 54.6\perfup{+4.9} & 55.4\perfup{+1.2} \\
    threshold=1.0 & 53.4\perfup{+1.1} & 54.0\perfup{+4.3} & 55.1\perfup{+0.9} \\
    soft-weight   & 56.8\perfup{+4.5} & 56.9\perfup{+7.6} & 56.6\perfup{+2.4} \\
    \bottomrule
    \end{tabular}
    }
}
\vspace{-.6cm}
\end{table}

\vspace{-10pt}
\subsection{Limitations}
\vspace{-5pt}
One limitation of our proposed data selection method is that it requires specifying some hyperparameters (\eg the parameters of the beta distribution). %
Notably, we use the same hyperparameters across all datasets and demonstrate the robustness of our method to different choices. 
There is also an additional cost incurred by our approach associated with computing the source data training weights. 
However, this is negligible compared to the overall cost of training. 
Like all discovery methods, if the resulting discovered categories were not verified by a human expert, there could be potential negative impact of our work if it was applied in safety-critical applications.

\vspace{-10pt}
\section{Conclusion}
\label{sec:conclusion}
\vspace{-10pt}
We explored the problem of category discovery through the lens of labeled data selection. 
Perhaps surprisingly, we showed that the precise choice of labeled data matters significantly for discovery performance. 
We demonstrated that the most effective labeled source is not too dissimilar to the unlabeled data as to provide no useful signal, nor is it too similar as to cause confusion between novel and  previously seen similar categories. 
We presented a new approach for automatically weighting different instances from the labeled dataset which results in improved discovery performance compared to standard splits from the existing literature. 
Interestingly, we showed that this improvement is not the result of simply selecting more labeled data, but in fact less, \ie less is more.  
We hope that our work opens up a new line of enquiry into data selection for category discovery.

\vspace{5pt}
\noindent {\bf Acknowledgements.}  This work was in part supported by a Royal Society Research Grant, the Edinburgh–Copenhagen Strategic Partnership Seed-Fund, and the Pioneer Centre for AI, DNRF grant number P1.

\bibliographystyle{splncs04}
\bibliography{main}

\clearpage
\appendix
\setcounter{table}{0}
\renewcommand{\thetable}{A\arabic{table}}
\setcounter{figure}{0}
\renewcommand{\thefigure}{A\arabic{figure}}
\noindent{\LARGE \bf Appendix}

\section{Soft weighting - Beta method}
In this section, we aim to provide additional ablation results to justify the design choices we made in the main text.

\subsection{Additional ablations} 

\noindent\textbf{Ablation on $\alpha$ and $\beta$ values.} In~\cref{tab:alpha_beta}, we present more results (\ie in additional to the results in Tab.~\textcolor{red}{5} in the main paper) where we use a  wider range of $\alpha$ and $\beta$ values.  
We observe that the default values of $\alpha=\beta=5$ across all datasets offer a competitive performance among other choices, and that when we chose parameters that select labeled examples that are more similar to the unlabelled set (\ie $\alpha>\beta$) results in worse performance. 

\noindent\textbf{Ablation of using reweighting \vs resampling.}
The weight we obtained from Eq.13 from the main paper can also be used to resample the dataset rather than weighting the loss value.
The resampling of the dataset can be done by normalizing the weight associated with each of the categories to a probability value and then using this probability to sample data from the dataset to form training mini-batches. 
When using resampling, we use equal weighting in the loss function during training.
In~\cref{tab:threshold_cut}, we present the comparison of using the weight for resampling the dataset and weighting the loss.
The difference between the two approaches is marginal, thus we choose to default the design of the beta method to do soft weighting on the losses.

\noindent\textbf{Ablation of the similarity metric.}
In Eq. 12 of the main paper, we compute the similarity score as the cosine similarity of the labeled centroids to the farthest unlabeled data point. 
However, other ways of computing this similarity are also potentially valid.  
In~\cref{tab:design_choice_eq12}, we evaluate using other choices, such as using the maximum similarity of the labeled centroids to the unlabeled data and the median of the all cosine similarity score between the labeled centroids to the unlabeled data points.
We can see from the results that using the cosine similarity of the most similar unlabeled data (\ie `max') leads to a performance drop compared to the no selection baseline, while using the median value of all similarities and the minimum value of the similarities yield similar results.

\begin{table}[h]
\centering
\parbox{.5\linewidth}{
    \centering
    \caption{Ablations on more values of $\alpha$ and $\beta$. We default to $\alpha=\beta=5$ for our experiments. Scores are performance on `New' categories.}
    \label{tab:alpha_beta}
    \vspace{-.4cm}
    \resizebox{0.45\columnwidth}{!}{
    \begin{tabular}{cc| lll}
    \toprule
    $\alpha$  & $\beta$    &  CUB  &  Aircraft  & SDogs \\
    \midrule
    1 & 1 &   52.3    &   49.3     & 54.2 \\
    7 & 3 &   42.4\perfdown{-9.9}  &   38.5\perfdown{-10.8}      & 46.0\perfdown{-7.2} \\
    9 & 3 &   40.2\perfdown{-12.1} &   36.7\perfdown{-12.6}     & 45.2\perfdown{-9.0}\\
    3 & 7 &   57.4\perfup{+5.1}    &   55.4\perfup{+6.1}     & 57.2\perfup{+3.0}\\
    3 & 9 &   57.7\perfup{+5.4}    &   55.8\perfup{+6.5}     & 56.1\perfup{+1.9}\\
    5 & 5 &   56.8\perfup{+4.5}    &   56.9\perfup{+7.6}     & 56.6\perfup{+2.4} \\
    \bottomrule
    \end{tabular}
    }
}
\hfill
\parbox{.48\linewidth}{
    \centering
    \caption{%
    Our main experiments use soft weighting. Here we ablate using the weight for resampling the dataset instead of weighting the loss. Scores are performance on `New' categories.
    }
    \label{tab:threshold_cut}
    \vspace{-.4cm}
    \resizebox{0.5\columnwidth}{!}{
    \begin{tabular}{l| lll}
    \toprule
      &  CUB  &  Aircraft  & SDogs \\
    \midrule
    baseline      & 52.3    &   49.3     & 54.2 \\
    resampling   & 56.5\perfup{+4.2} & 57.2\perfup{+7.9} & 56.9\perfup{+2.7} \\
    soft-weight   & 56.8\perfup{+4.5} & 56.9\perfup{+7.6} & 56.6\perfup{+2.4} \\
    \bottomrule
    \end{tabular}
    }
}
\end{table}

\begin{table}[h]
    \centering
    \caption{Ablations on the design choice of Eq. 12 in the main paper. We default to using the minimum similarity of the labeled centroids to the unlabeled data.}
    \label{tab:design_choice_eq12}
    \vspace{-.4cm}
    \begin{tabular}{l| lll}
    \toprule
      &  CUB  &  Aircraft  & SDogs \\
    \midrule
    baseline & 52.3    &   49.3     & 54.2 \\
    max    & 47.9\perfdown{-4.4} & 45.1\perfdown{-4.2} & 50.2\perfdown{-4.0} \\
    median & 57.0\perfup{+4.7} & 56.7\perfup{+7.4} & 56.2\perfup{+2.0} \\
    min    & 56.8\perfup{+4.5}   & 56.9\perfup{+7.6}   & 56.6\perfup{+2.4} \\
    \bottomrule
    \end{tabular}
\end{table}

\section{Binning method}

\subsection{Implementation details}
\begin{figure}
    \centering
    \includegraphics[width=\linewidth]{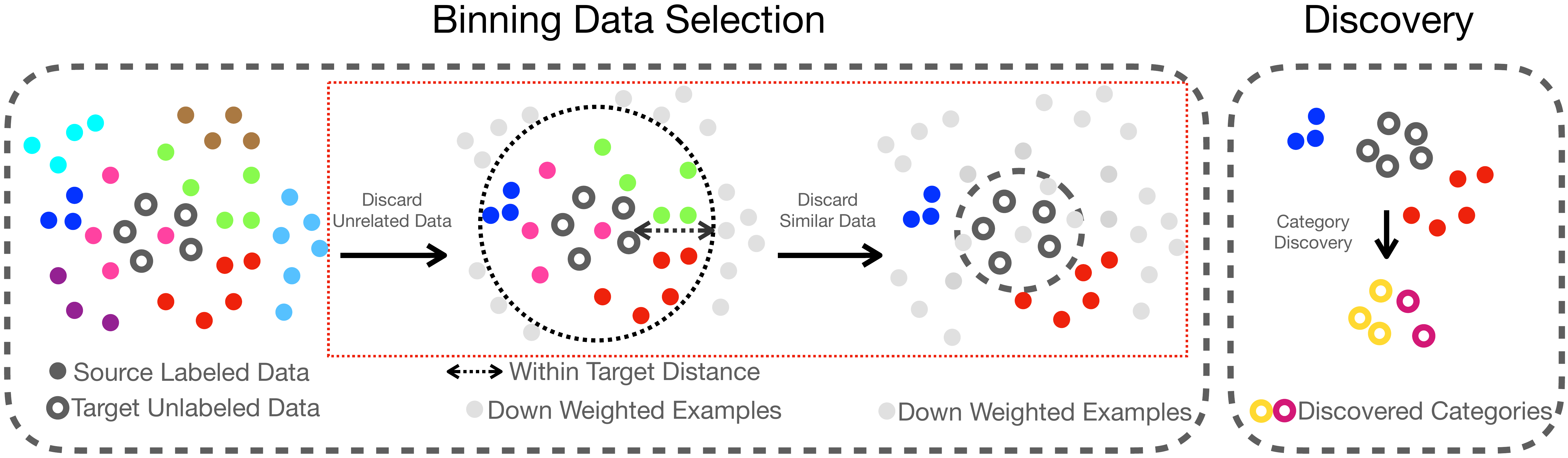}
    \caption{Overview of our binning labeled data selection process for category discovery. We first discard labeled source data based on a threshold calculated from within the unlabeled target data. 
    Next we discard data that is too similar to the unlabeled target data. The remaining labeled source data, along with the unlabeled target data, is then fed to a category discovery method. }
    \label{fig:binning_method}
\end{figure}

In the main text we briefly outlined our hard selection-based binning method. 
In this section, we provide more details on how the binning method is implemented.
The binning method for data selection simply `chunks' the labeled source dataset into several equal-sized subsets based on the similarity to the target unlabeled set, and then selects the chunks that are not too similar or too dissimilar. 
The difficulty is that we cannot know how unrelated data is in the labeled data pool, thus we cannot determine how many chunks to use or which chunk to select. 
To address this, we design a selection method that first filters out unrelated data and then performs selection after. Our binning approach is illustrated in~\cref{fig:binning_method}. 

To successfully filter out unrelated data from the source, we would need a distance threshold.
Setting a fixed value for the threshold would be hard, requiring a lot of trial and error. 
Thus we devised a method for automatically obtaining it. 
The high-level idea is that we can compute the distances between the data points \emph{within} the target unlabeled data. 
This should give us a measure of how distant the categories in the target unlabeled data should be, and thus we can filter out data that has a larger distance than this threshold.
To do this, we first randomly divide the target unlabeled data $\mathcal{D}^u$ into two sets, $\mathcal{D}_0^u$ and $\mathcal{D}_1^u$, each containing the same number of estimated clusters. 
Then with these sets we can compute: 
\begin{equation}
    \text{EMD}(\mathcal{D}^l \cup \mathcal{D}_0^u, \mathcal{D}_1^u) = \frac{\sum_{i\in I(\mathcal{D}^l\cup \mathcal{D}_0^u)} \sum_{j\in I(\mathcal{D}_1^u)} k_{i,j} d_{i,j}}{\sum_{i\in I(\mathcal{D}^l\cup \mathcal{D}_0^u)} \sum_{j\in I(\mathcal{D}_1^u)} k_{i,j}},
    \label{eq:EMD_sel}
\end{equation} 
where $I(\mathcal{D})$ stands for the indexes of the mean feature vectors of $\mathcal{D}$, $d_{i,j}$  is the distance between  mean feature vectors, and $k_{i,j}$ is optimal flow. 

\noindent\textbf{Discarding distant unrelated data.}
$\mathcal{D}_0^u$ is used during the  Earth Mover’s Distance (EMD) computation to obtain a threshold value that can be used to filter out distant unrelated labeled categories that are dissimilar to the target data. 
We start by defining:
\begin{equation}
    d_{i,:}=\frac{\sum_{j\in I(\mathcal{D}_1^u)} k_{i,j}d_{i,j}}{\sum_{j\in I(\mathcal{D}_1^u)} k_{i,j}},
\end{equation}
where $i\in I(\mathcal{D}^l\cup \mathcal{D}_0^u)$, $d_{i,:}$ denotes the distance of one source category/cluster $i$ to the target unlabeled data, and $k_{i,j}$ is the flow from \cref{eq:EMD_sel}. 
By calculating $\bar{d}_{\mathcal{D}^u}=\frac{1}{|I(\mathcal{D}_0^u)|}\sum_{i\in I(\mathcal{D}_0^u)} d_{i,:}$, which can be understood as the distance within the target unlabeled data itself, we can filter out any entry in $I(\mathcal{D}^l)$ that is in  $\{i| d_{i,:} > \bar{d}_{\mathcal{D}^u}, i\in I(\mathcal{D}^l)\}$. 
The rationale behind this is that if a labeled category has a distance larger than $\bar{d}_{\mathcal{D}^u}$, it would be too distant from the target unlabeled data to provide a useful learning signal for a category discovery method. 
After this filtering, the remaining dataset $\mathcal{D}^r$  will contain labeled categories that have a smaller distance than the threshold $\bar{d}_{\mathcal{D}^u}$.

\noindent\textbf{Selecting relevant data.} 
In the previous step, we filtered out data from the source that was too dissimilar to the target. 
Next, we need to remove data that is too \emph{similar}. 
To do this, we rank the labeled categories in the remaining source data $\mathcal{D}^r$ based on their distance to the target data. 
We then divide $\mathcal{D}^r$ into $L$ equal-sized subsets $\mathcal{D}^r=\bigcup_{l=1}^L \mathcal{D}^r_l$, where $\mathcal{D}^r_1$ contains the most similar labeled categories to the target data, and $\mathcal{D}^r_L$ contains the most dissimilar labeled categories in $\mathcal{D}^r$.
Empirically, we found that selecting the distant subsets such as $\mathcal{D}^r_L$ gives a better performance compared to selecting the similar ones such as $\mathcal{D}^r_1$. 
Perhaps surprisingly, we later show that the performance of using $\mathcal{D}^r_L$ sometimes outperforms using the entire subset $\mathcal{D}^r$, and can even outperform models that use all the original labeled data $\mathcal{D}^l$.

In practice, we randomly split $\mathcal{D}^u$ into $\mathcal{D}^u_0$ and $\mathcal{D}^u_1$ multiple times to get a more robust estimate of $\mathcal{D}^r$. 
We then select $\mathcal{D}^r_L$ as our new labeled dataset for supervising the category discovery process, where $L=2$. 
We choose $L=2$, as it strikes a balance of not being too similar, but still relevant to the task. 
The selection is done by assigning the weight of $1$ to examples in $\mathcal{D}^r_L$ and assigning a weight of $0$ to all the other examples. 
For implementation, as the examples with $0$ weight will not influence the training process, we can discard them and only sample from $\mathcal{D}^r_l$ for training.

\subsection{Ablations of the binning method}

In this section, we provide ablations of the binning data selection method.  
The evaluation is performed using the CUB~\cite{cub200}, Stanford Dogs, and FGVC-Aircraft~\cite{aircraft} datasets from SSB~\cite{vaze2021open}. 
The selection of labeled data is done using the combination of SSB, Stanford Dogs, and iNat-Insect as the source set, and we report the clustering accuracy on the `New' categories.

\noindent{\bf Directly partition the entire labeled source.}
Our proposed binning selection method has two steps. 
The first is to obtain a threshold $\bar{d}_{\mathcal{D}^u}$ for discarding irrelevant data from the labeled source. 
Then we chunk the remaining data into two equal-sized subsets based on the distance to the target data. 
The second subset is then selected for supervising the category discovery models. 
In~\cref{tab:direct_split}, we present the results of ablating these two parts.
We can see that if we remove the `discard unrelated data' step (\ie Ours w/o Discard), the performance drops significantly.  This is because we end up selecting unrelated data to the target  for supervising the category discovery model.
Next, we remove the second selection step, and instead directly use all remaining data after the removal of unrelated data. 
Again, the performance also drops, as similar data is selected and the model gets confused.

\begin{table}[t]
\parbox{.45\linewidth}{
    \centering
     \caption{Here we ablate the major parts of the binning data selection method. 
    We report accuracy on `New' categories and use the `all dataset' as the source pool. 
    `Ours' stands for applying the binning method on SimGCD.
    }
    \vspace{-8pt}
    \resizebox{0.45\columnwidth}{!}{
    \begin{tabular}{l ccc}
    \toprule
         &  CUB  &  Aircraft  & SDogs \\
    \midrule
    Ours w/o Discard   &   12.1    &   8.9      &  10.0\\
    Ours w/o Selection &   45.7    &   41.4      & 47.2 \\
    Ours               &   56.7    &   56.8      &  55.6 \\
    \bottomrule
    \end{tabular}
    }

    \label{tab:direct_split}
}
\hfill
\parbox{.45\linewidth}{
    \centering
        \caption{Here we estimate the number of clusters using off-the-shelf estimation methods. We report accuracy on `New' categories where models use the `all dataset' source pool. 
    `Ours' stands for applying the binning method on SimGCD.
    The final row of this table is using the ground truth number of categories.
    }
    \vspace{-8pt}
    \resizebox{0.38\columnwidth}{!}{
    \begin{tabular}{l ccc}
    \toprule
         &  CUB    &   Aircraft     &   SDogs\\
    \midrule
    Ours w/ \cite{vaze2022generalized} &   55.7    &   53.7      & 54.7\\
    Ours w/ \cite{zhao2023learning}    &   54.2    &   52.9      & 53.1    \\
    Ours w/ GT                         &   56.7    &   56.8      & 55.6 \\
    \bottomrule
    \end{tabular}
    }
    \label{tab:number}
}
\vspace{-8pt}
\end{table}

\noindent{\bf Using estimated category numbers for clustering.}
To compute the mean feature vectors $\bar{\mathbf{h}}_c^u$ on the unlabeled target data we need to have the knowledge of how many clusters to use for a clustering algorithm like $k$-means. 
The common assumption of category discovery models is that the number of clusters in the target data is known \textit{a priori}~\cite{vaze2022generalized,wen2023parametric,fei2022xcon}. 
Each of our baseline methods also receives this information.
In this section, we study the performance of our binning selection method when the number of clusters is unknown and has to be estimated using off-the-shelf cluster number estimation methods~\cite{zhao2023learning,vaze2022generalized}. 
In~\cref{tab:number}, we present a performance comparison using an estimated number of clusters.
Our binning selection method is robust to the number of clusters used.

\subsubsection{Sensitivity to hyperparameters.}
Here we provide additional ablations on the impact of the hyperparameters of the binning method.
Specifically, we investigate the role of the number of chunks $L$, the distance metric used, and the number of clusters.

\cref{tab:ablation_L} presents the ablation of the value of $L$. 
For the results in the main paper we select $L=2$. 
For the different values of $L$ explored in this ablation, we always select the $L$-th chunk for supervising the model, \ie if $L=3$, we discard the first two and use the 3rd chunk for supervising the category discovery model. 
We can see that our binning method is robust to a range of $L$, and note that higher values of $L$ will result in smaller numbers of images per chunk, thus when the value of $L$ is higher, the performance starts to decrease.

\begin{table}[h]
    \centering
    \caption{Ablation of the number of chunks $L$.
    `Ours' stands for applying the binning method on SimGCD.
    }
    \vspace{-5pt}
    \resizebox{0.6\linewidth}{!} 
    {
    \begin{tabular}{l lll lll}
    \toprule
    Method     & \multicolumn{3}{c}{CUB}  & \multicolumn{3}{c}{FGVC-Aircraft}\\
    \cmidrule(lr){2-4}\cmidrule(lr){5-7}
               &  All & Old & New  & All & Old & New \\
    \midrule
    Ours w. $L=2$       & 58.2 & 64.6 & 56.7 & 55.7 & 57.9 & 56.8\\
    Ours w. $L=3$       & 56.1 & 62.3 & 55.0 & 53.4 & 55.2 & 53.4\\
    Ours w. $L=4$       & 57.1 & 62.0 & 55.6 & 56.1 & 58.2 & 55.7\\
    Ours w. $L=5$       & 53.0 & 57.8 & 50.1 & 48.7 & 51.4 & 49.1\\
    \bottomrule
    \end{tabular}
    }
    \vspace{-10pt}
    \label{tab:ablation_L}
\end{table}

\begin{table}[h]
    \centering
    \caption{Ablation of distance metric used. 
    The results here are based on applying binning on SimGCD.
    }
    \vspace{-5pt}
    \resizebox{0.55\linewidth}{!}{
    \begin{tabular}{l lll lll}
    \toprule
    Method     & \multicolumn{3}{c}{CUB}  & \multicolumn{3}{c}{FGVC-Aircraft}\\
    \cmidrule(lr){2-4}\cmidrule(lr){5-7}
               &  All & Old & New  & All & Old & New \\
    \midrule
    Euclidean       &   58.2 & 64.6 & 56.7 & 55.7 & 57.9 & 56.8\\
    Cosine          &   57.2 & 64.0 & 56.4 & 55.4 & 57.3 & 56.2\\
    $\ell_2$ Norm   &   59.0 & 64.7 & 56.9 & 55.8 & 57.4 & 57.2\\
    \bottomrule
    \end{tabular}
    }
    \vspace{-10pt}
    \label{tab:dist_metric}
\end{table}

\cref{tab:dist_metric} presents the ablation on different distance metrics to use when calculating the Earth Mover's Distance (EMD). 
The Euclidean distance is used by default in the main paper. 
The other distance metrics we tested include the cosine distance and an alternative where we first $\ell_2$ normalize the features and then compute the Euclidean distance.
We are robust to these choices, and we can see that the performance is slightly better when using the $\ell_2$ normalized Euclidean distance.

\begin{table}[h]
    \centering
    \caption{Ablation where we vary the number of clusters. 
    The results are based on applying binning on SimGCD.
    }
    \vspace{-5pt}
    \resizebox{0.55\linewidth}{!}{
    \begin{tabular}{l lll lll}
    \toprule
    Method     & \multicolumn{3}{c}{CUB}  & \multicolumn{3}{c}{FGVC-Aircraft}\\
    \cmidrule(lr){2-4}\cmidrule(lr){5-7}
               &  All & Old & New  & All & Old & New \\
    \midrule
    GT                        & 58.2 & 64.6 & 56.7 & 55.7 & 57.9 & 56.8\\
    $K=50$                    & 55.4 & 60.3 & 53.1 & 54.2 & 57.7 & 55.8\\
    $K=100$                   & 57.2 & 63.5 & 56.4 & 53.2 & 56.2 & 53.1\\
    $K=500$                   & 52.3 & 57.2 & 49.3 & 50.1 & 51.2 & 47.6\\
    $K=1000$                  & 49.6 & 54.2 & 46.2 & 45.7 & 48.9 & 44.3\\
    \bottomrule
    \end{tabular}
    }
    \vspace{-10pt}
    \label{tab:number_clusters}
\end{table}

Another important ablation is the number of clusters to use when clustering the unlabeled data.
Note that all the baseline category discovery methods we have used in the main paper assume the number of clusters is known. 
As noted earlier, this is a common assumption in the category discovery literature.
In \cref{tab:number}, we presented the results of using different cluster number estimation methods for our binning selection methods. 
In~\cref{tab:number_clusters} we present the results of setting the number of clusters to a fixed value.
We can see that when the fixed number is close to the true number of clusters (\ie $100$ for CUB and $50$ for Aircraft), the model gets the best performance. 
When the number is larger, the performance degrades gradually.

\section{Datasets}
In~\cref{tab:dataset}, we present the dataset statistics for datasets we used in our experiments.

\begin{table}[h]
    \centering
    \caption{
    The statistics of the datasets we used for our experiments.
    }
    \vspace{-5pt}
    \resizebox{0.65\linewidth}{!}{
    \begin{tabular}{lrcrc}
    \toprule
                & \multicolumn{2}{c}{Labeled}  & \multicolumn{2}{c}{Unlabeled}\\
                \cmidrule(rl){2-3}\cmidrule(rl){4-5}
    Dataset           & \#Image   & \#Class   & \#Image   & \#Class \\
    \midrule
    CUB~\cite{cub200}             & 1.5K      & 100       & 4.5K      & 200 \\
    Stanford Cars~\cite{stanfordcars}   & 2.0K      & 98        & 6.1K      & 196 \\    
    FGVC-Aircraft~\cite{aircraft}   & 1.7K      & 50        & 5.0K      & 50 \\
    NABirds~\cite{van2015building}  &   12K        &     200       &    36K        &    400    \\
    iNat-Insect~\cite{van2018inaturalist}  &   31.5K    &     1263       &     94.8K    &   2526       \\
    Stanford Dogs~\cite{stanforddogs}    &    3K     &     60     &    9K      &    120    \\
    Herbarium-19~\cite{tan2019herbarium}    & 8.9K      & 341       & 25.4K     & 683 \\
    ImageNet-100~\cite{russakovsky2015imagenet}    & 31.9K     & 50        & 95.3K     & 100 \\
    ImageNet-1k-SSB~\cite{russakovsky2015imagenet}    & 1284K     & 2000        & 1920K     & 2000 \\
    \bottomrule
    \end{tabular}
    }
    \vspace{-10pt}
    \label{tab:dataset}
\end{table}

\section{Discovery methods}
For completeness, in this section we briefly describe the GCD~\cite{vaze2022generalized}, XCon~\cite{fei2022xcon}, and $\mu$GCD~\cite{vaze2023no} methods that we have used for some of the comparisons in the main paper.  
GCD and XCon focus on the representation learning for the GCD tasks, while the classifier is directly implemented as a semi-supervised $k$-means classifier. 
$\mu$GCD utilizes loss for both representation learning and classifier learning, with the classifier implemented similarly to the parametric classifier in SimGCD.

For all four methods we evaluated (including SimGCD from the main paper), we assume that the number of categories in the unlabeled dataset is known. 
In practice where the number of categories is not known, it can be estimated using off-the-shelf category number estimation methods like semi-supervised $k$-means~\cite{vaze2022generalized}.
In~\cref{tab:number,tab:number_clusters}, we present the results of using off-the-shelf categories number estimation methods and the results of using fixed number of categories.

\subsubsection{GCD.}
For representation learning, GCD~\cite{vaze2022generalized} utilizes contrastive learning. 
Specifically, for the unlabeled data $\mathbf{x}_i\in B$, a self-supervised contrastive loss is employed: 
\begin{equation}
    \mathcal{L}_{\text{rep}}^{u}=\frac{1}{|B|} \sum_{\mathbf{x}_i\in B} - \log \frac{\exp (\hat{\mathbf{z}}_i^\top \tilde{\mathbf{z}}_i / \tau_u)}{\sum_{\mathbf{x}_j \in B} \exp (\hat{\mathbf{z}}_j^\top \tilde{\mathbf{z}}_i / \tau_u)},
\end{equation}
where $\mathbf{z}_i =m(f(\mathbf{x}_i))$ is the projected feature for contrastive learning, $\hat{\mathbf{z}}_i$ and $\tilde{\mathbf{z}}_i$ are the features of two augmentations of the same image $\mathbf{x}_i$, $\tau_u$ is a temperature value, and $B$ is a mini-batch of unlabeled images.
For the labeled data $\mathbf{x}_i\in B^l$, a supervised contrastive loss is employed to learn a discriminative representation:
\begin{equation}
    \mathcal{L}_{\text{rep}}^{s} = \frac{1}{|B^l|}\sum_{\mathbf{x}_i\in B^l}\omega_i\frac{1}{|\mathcal{N}_i|}\sum_{p\in\mathcal{N}_i}-\log\frac{\exp (\hat{\mathbf{z}}_i^\top \tilde{\mathbf{z}}_p / \tau_s)}{\sum_{n\neq i}\exp (\hat{\mathbf{z}}_i^\top \tilde{\mathbf{z}}_n / \tau_s)},
\end{equation}
where the indices of images with the same label as $\mathbf{x}_i$ is stored in $\mathcal{N}_i$, $\omega_i$ is the weight for example $i$, and $\tau_s$ is another temperature value.
The representation used for category discovery is learned using these two losses with a weighting factor $\lambda$, $\mathcal{L}_{\text{rep}}=(1-\lambda) \mathcal{L}_{\text{rep}}^u + \lambda \mathcal{L}_{\text{rep}}^s$.

After the representation is learned, a semi-supervised $k$-means algorithm is applied for assigning labels to the unlabeled data.
This algorithm is modified from the original unconstrained $k$-means algorithm by forcing the assignment of the labeled data to be the ground truth label.
Specifically, this algorithm first obtains the initial centroids using $k$-means++~\cite{arthur2007k}, updates the centroids,  and then assigns labels like the $k$-means algorithm except it forces the labels of the labeled data to be the ground truth.

\subsubsection{XCon.}
XCon~\cite{fei2022xcon} extends GCD by proposing dataset partitioning to learn more discriminative representations.
Specifically, it first partitions the whole dataset $\mathcal{D}=\{(\mathbf{x}_i, y_i)\}$ into $K$ sub-dataset $\{\mathcal{D}_1, \mathcal{D}_2, \dots, \mathcal{D}_K\}$ using a $k$-means clustering of a self-supervised representation, and then uses the same supervised and self-supervised contrastive losses as GCD on each of the sub-datasets (\ie by sampling negative and positive samples only within the sub-dataset). 
XCon claims that this partitioning can help the sampling of hard negative examples, and thus help representation learning.
Apart from these contrastive learning losses on the partitioned dataset, XCon also employs the same contrastive losses over the whole dataset as GCD.
The label assignment is performed similarly to GCD, using a semi-supervised $k$-means algorithm.

\subsubsection{$\mu$GCD.}
$\mu$GCD~\cite{vaze2023no} extends SimGCD~\cite{wen2023parametric} method, which we introduced in the main paper.
$\mu$GCD adopts a `cosine' classifier~\cite{gidaris2018dynamic} with the features being `L2'-normalized to help the model learn a more balanced classifier across old and new categories.
The loss functions for learning are the same as the SimGCD learning losses.
Different from SimGCD, $\mu$GCD makes use of a teacher-student architecture, where the student is fed with a strongly augmented image and is trained with gradient descent, and the teacher uses a weakly augmented image and is updated via moving average from the student model.

\end{document}